%% file: main.tex
\documentclass[10pt,twocolumn,letterpaper]{article}

\usepackage[pagenumbers]{cvpr} 

\definecolor{cvprblue}{rgb}{0.21,0.49,0.74}
\usepackage[pagebackref,breaklinks,colorlinks,allcolors=cvprblue]{hyperref}

\usepackage[utf8]{inputenc}
\usepackage{algorithm}
\usepackage{algpseudocode}
\usepackage{adjustbox}

\def\confName{CVPR}
\def\confYear{2025}

\usepackage[pagenumbers]{cvpr} 


\definecolor{cvprblue}{rgb}{0.21,0.49,0.74}
\usepackage[pagebackref,breaklinks,colorlinks,allcolors=cvprblue]{hyperref}

\usepackage[utf8]{inputenc}
\usepackage{algorithm}
\usepackage{algpseudocode}
\usepackage{adjustbox}

\def\confName{CVPR}
\def\confYear{2025}

\usepackage{hyperref}

\usepackage{orcidlink}
\usepackage{booktabs} 
\usepackage{multirow}
\usepackage{xcolor}
\usepackage{colortbl}
\usepackage{booktabs}   
\usepackage{multirow}   
\usepackage{graphicx}   
\newcommand{\name}{FineEdit}

\begin{document}

\title{FineEdit: Fine-Grained Image Edit with Bounding Box Guidance} 


\author{
Haohang Xu \quad 
Lin Liu \quad 
Zhibo Zhang \quad 
Rong Cong \quad 
Xiaopeng Zhang \quad 
Qi Tian  \\
 Huawei Inc. \;
}

\twocolumn[{%
\renewcommand\twocolumn[1][]{#1}%
\maketitle
\begin{center}
    \centering
    \includegraphics[width=1.0\linewidth]{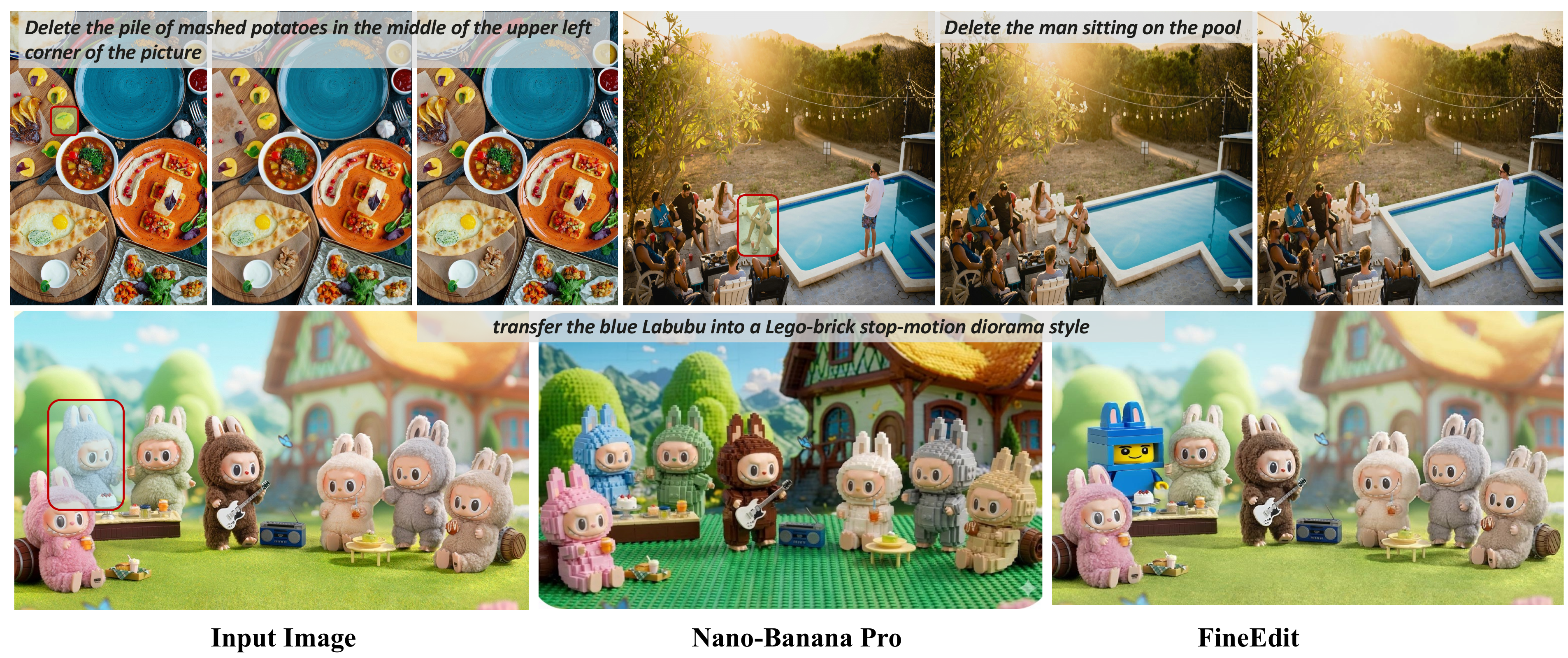}
    \vspace{-4mm} 
    \captionof{figure}{Comparison between the SOTA closed source model Nano-Banana Pro, and our proposed \name. The results (left: Input; middle: Nano-Banana Pro; right: \name) demonstrate that even the Nano family models struggle with precise object localization in complex scenarios. In contrast, leveraging bounding-box instructions, \name\ achieves superior editing fidelity and localization accuracy.}
    \label{fig:motivation}
\end{center}%
}]



\begin{abstract}
Diffusion-based image editing models have achieved significant progress in real world applications. However, conventional models typically rely on natural language prompts, which often lack the precision required to localize target objects. Consequently, these models struggle to maintain background consistency due to their global image regeneration paradigm. Recognizing that visual cues provide an intuitive means for users to highlight specific areas of interest, we utilize bounding boxes as guidance to explicitly define the editing target. This approach ensures that the diffusion model can accurately localize the target while preserving background consistency. To achieve this, we propose \textbf{FineEdit}, a multi-level bounding box injection method that enables the model to utilize spatial conditions more effectively. To support this high precision guidance, we present \textbf{FineEdit-1.2M}, a large scale, fine-grained dataset comprising 1.2 million image editing pairs with precise bounding box annotations. Furthermore, we construct a comprehensive benchmark, termed \textbf{FineEdit-Bench}, which includes 1,000 images across 10 subjects to effectively evaluate region based editing capabilities. Evaluations on FineEdit-Bench demonstrate that our model significantly outperforms state-of-the-art open-source models (e.g., Qwen-Image-Edit and LongCat-Image-Edit) in instruction compliance and background preservation. Further assessments on open benchmarks (GEdit and ImgEdit Bench) confirm its superior generalization and robustness.
\end{abstract}

\section{Introduction}
\label{sec:intro}
Diffusion models have recently achieved substantial advancements across various generative applications, including image generation \cite{ho2020denoising,song2020score,saharia2022photorealistic,rombach2022high}, video generation \cite{singer2022make,blattmann2023stable,wan2025wan}, and their respective editing tasks. Fundamentally, these models transform random noise into the target data distribution via a progressive denoising process. State-of-the-art diffusion-based image editing models \cite{labs2025flux,wu2025qwen,seedream2025seedream} now produce high quality results that rival professional human editing. However, conventional methods \cite{meng2021sdedit, brooks2023instructpix2pix, labs2025flux} typically rely exclusively on text-based instructions to specify the location and goal of an editing task. While effective in simple, object centric scenarios, these approaches often falter in complex, real world scenes containing multiple objects. As shown in Fig. \ref{fig:motivation}, in complex multi-object scenarios, text alone is often insufficient to unambiguously specify the target. This limitation introduces two primary shortcomings: first, the model may fail to precisely localize the target object, resulting in inaccurate editing; and second, the absence of explicit spatial constraints leaves the background vulnerable to unintended changes, such as shifts in color scheme or layout.

Recent research has extensively explored the incorporation of visual priors into diffusion models to enhance controllability. For instance, ControlNet \cite{zhang2023adding} utilizes side networks to inject structural guidance. However, it typically operates at a holistic level, lacking the precise localization capabilities required for complex, multi-object scenes. Alternatively, inpainting-based approaches leverage masks to isolate specific regions. Within this paradigm, training-free methods like RePaint \cite{lugmayr2022repaint} are constrained by the generative limits of the frozen base model, while training-based models such as BrushNet \cite{ju2024brushnet} are primarily optimized for content restoration (i.e., recovering missing pixels) rather than semantic modification. More recently, state-of-the-art editing models (e.g., Qwen-Image-Edit \cite{wu2025qwen} and LongCat-Image-Edit \cite{LongCat-Image}) have allowed users to define editing areas via bounding boxes. Nevertheless, these methods often yield unsatisfactory results. Typical issues include boundary leakage, where edits unintentionally spill into the surrounding background, and visual artifacts, such as residual traces of the bounding boxes appearing in the final output (illustrated in the appendix). Consequently, existing approaches still struggle to achieve precise, artifact-free editing within strictly defined regions. 
%

%

%

To address these issues, we introduce \textbf{FineEdit}, a novel framework that leverages bounding-box visual priors to explicitly guide the image editing process. FineEdit resolves two critical challenges: First, we propose a multi-level visual instruction injection architecture that effectively utilizes spatial constraints to guide the denoising process. To further align the model with human perception, we introduce a decoupled post-training reinforcement learning strategy. This strategy independently evaluates foreground modifications and background consistency, overcoming the misaligned reward signals inherent in conventional global VLM scoring. Second, to support the high-precision training paradigm, we present \textbf{FineEdit-1.2M}, a rigorously filtered, high quality dataset comprising 1.2 million image editing pairs. Unlike previous editing datasets \cite{zhao2024ultraedit,yu2025anyedit,ge2024seed,Layer2025NoHumansRequired,wang2025gptimageedit15mm,wei2024omniedit}, FineEdit-1.2M ensures high editing quality and strict background consistency. Extensive experiments demonstrate that FineEdit significantly improves spatial instruction following capabilities and background preservation, outperforming state-of-the-art editing models by a large margin.

The key contributions of are summarized as follows:

\vspace{1mm}
$\bullet$ We propose \name, a novel architecture utilizing multi-level visual instruction injection to guide the denoising process via bounding-box priors. Additionally, we introduce a decoupled post-training reinforcement learning strategy that independently evaluates foreground edits and background preservation, resolving the reward misalignment typical of global VLM scoring

\vspace{1mm}
$\bullet$ We construct FineEdit-1.2M, a large scale dataset containing 1.2 million high quality image editing pairs with precise spatial annotations and strict consistency.

\vspace{1mm}
$\bullet$ Evaluations demonstrate that \name \ significantly surpasses existing editing models in both spatial instruction adherence and background preservation, establishing a new standard for fine-grained image editing.

\section{Related Work}
\label{sec:related}

\subsection{Diffusion-based Image Editing} 
Diffusion-based models \cite{rombach2022high} treat image editing as a conditional generation task, typically falling into two categories: training-free and training-based frameworks. Training-free methods \cite{hertz2022prompt,tumanyan2023plug,kulikov2025flowedit,cao_2023_masactrl,huberman2024edit,samuel2024lightning} manipulate the internal representations of a frozen model to preserve source structures (e.g., Prompt-to-Prompt \cite{hertz2022prompt}, Plug-and-Play \cite{tumanyan2023plug}). While bypassing fine-tuning costs, they are hyperparameter sensitive and struggle with complex structural modifications. Conversely, training-based models \cite{brooks2023instructpix2pix,zhang2023magicbrush,yu2025anyedit,zhao2024ultraedit,ge2024seed,Sheynin2023EmuEP} are fine-tuned to directly learn editing mappings, as pioneered by InstructPix2Pix \cite{brooks2023instructpix2pix}. However, most training-based methods rely solely on textual guidance, lacking the capacity to accept explicit visual spatial instructions (e.g., bounding boxes) for precise localized editing.

\subsection{Visual Instruction for Image Edit}
Recent works \cite{ma2023subject,zhao2024uni,huang2023composer,ye2023ipadapter,zhang2023adding,mou2024t2i} incorporate visual priors to enhance diffusion model controllability. Methods like ControlNet \cite{zhang2023adding}, T2I-Adapter \cite{mou2024t2i}, and GLIGEN \cite{li2023gligen} utilize side-networks or attention layers to accept structural maps or bounding boxes. However, these approaches are primarily optimized for conditional generation rather than instruction-based editing. Consequently, they exhibit two major limitations: First, they typically apply visual conditions globally, lacking the granularity to isolate specific objects for modification for background consistency. Second, they struggle to interpret edit instructions, often requiring complex engineering or additional inpainting masks to prevent unintended changes to the rest of the image.

\begin{figure*}[t]
    \centering
    \includegraphics[width=1.0\linewidth]{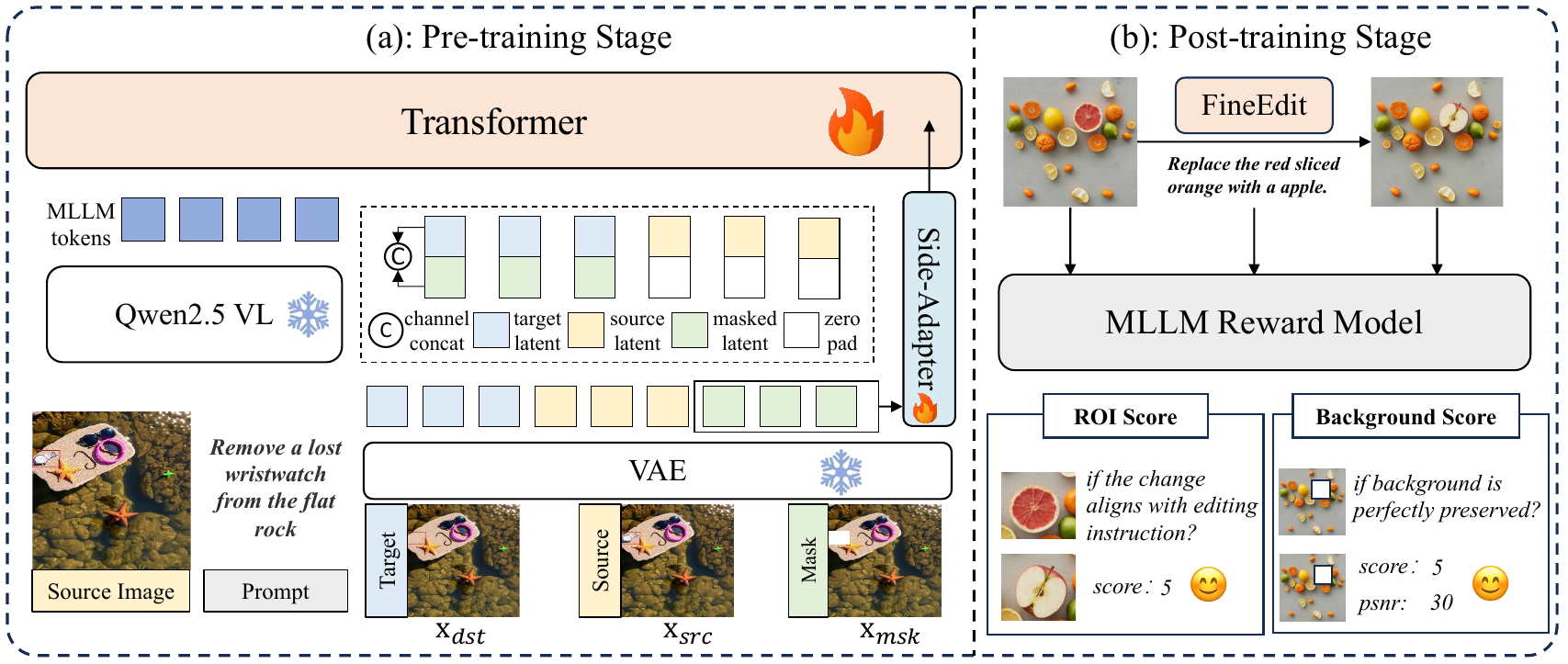}
        \vspace{-5mm}
    \caption{{Overview of \name\ framework, which includes two training stages: (a) Pre-training stage establishes multi-level spatial priors using early and deep fusion. (b) Post-training stage applies reinforcement learning with a novel decoupled reward function.}}
    \label{fig:arch_fineedit}
    \vspace{-4mm}    
\end{figure*}

Alternatively, text-guided inpainting methods \cite{fyp2025,ju2024brushnet,li2024brushedit,flux2024,manukyan2025hd,avrahami2022blended,xie2023smartbrush} enable localized generation within masked regions. However, these approaches are fundamentally designed for image reconstruction and completion rather than high level semantic editing. They prioritize seamless boundary transitions over complex structural transformations, often failing to follow intricate editing instructions. Furthermore, implementations relying on separately trained auxiliary modules (e.g., ControlNet-Inpaint \cite{zhang2023adding}) to handle mask constraints frequently suffer from suboptimal feature alignment. Consequently, they exhibit weaker preservation of unmasked regions and produce logically inconsistent edits compared to end-to-end trained editing models.
\section{Method}
\label{sec:method}

\subsection{An Overview of FineEdit Framework}
As illustrated in Fig. \ref{fig:arch_fineedit}, FineEdit builds upon the Qwen-Image \cite{wu2025qwen} architecture, employing Qwen2.5-VL \cite{bai2025qwen2} to encode conditional features from both the text prompt and the source image. The MM-DiT diffusion backbone accepts two primary inputs: noisy target image latents and clean source image latents. Both sets of latents are encoded via a frozen VAE \cite{kingma2013auto} and subsequently concatenated along the token sequence dimension.  The training process comprises two stages: \textit{Pre-training stage} establishes multi-level spatial priors using early and deep fusion mechanisms. In early fusion, VAE-encoded masked source tokens are concatenated with noisy target tokens along the channel dimension. In deep fusion, features extracted via a side-adapter are uniformly injected into intermediate transformer layers to enhance the perception of both input content and masked regions. \textit{Post-training stage} applies reinforcement learning with a novel decoupled reward function. By separating the conventional global reward into a foreground region-of-interest (ROI) score and a background preservation score, the framework achieves a highly precise assessment of localized editing quality. Each stage would be detailed below.


\subsection{Pre-training: Multi-Level Visual Instruction Injection}

\subsubsection{Preliminaries.} 
In the context of image editing, we are given a source image $\mathbf{x}_{src}$, a target image $\mathbf{x}_{dst}$, and a corresponding text prompt $c$. A pre-trained VAE first encodes $\mathbf{x}_{src}$ and $\mathbf{x}_{dst}$ into the latent space, yielding $\mathbf{q}_{src}$ and $\mathbf{q}_{dst}$, respectively. Subsequently, the forward diffusion process perturbs the target latent by adding noise: $\mathbf{\hat{q}}_{dst} = t\mathbf{\epsilon} + (1 - t)\mathbf{q}_{dst}$. Here, $\mathbf{\epsilon} \sim \mathcal{N}(\mathbf{0}, \mathbf{I})$ denotes Gaussian noise, and $t \in [0, 1]$ serves as the timestep parameter controlling the noise magnitude. 

Conventional image editing models~\cite{wu2025qwen, labs2025flux} typically concatenate the flattened source latent $\mathbf{q}_{src}$ and the noisy target latent $\mathbf{\hat{q}}_{dst}$ along the token sequence dimension. The combined sequence is input to the diffusion transformer $f$. In a flow matching framework, the model predicts the denoising velocity $\mathbf{v}$ by:
\begin{align}
\mathbf{v} = f([\mathbf{q}_{src}, \mathbf{\hat{q}}_{dst}] \mid t, c)
\end{align}

As aforementioned, such a pipeline lacks explicit instructions regarding spatial information, which yields two primary challenges: imprecise spatial localization and poor background consistency. To address these issues, we propose to fuse visual instructions via two distinct strategies: \textit{early fusion}, achieved by concatenating visual instructions at the input side, and \textit{deep fusion}, implemented by injecting visual instruction features into the deep transformer blocks.
\subsubsection{Early Fusion.} 
Visual instructions typically possess a sparse formulation, for instance, a bounding box is defined by only four parameters. To encode such instructions efficiently, we integrate them directly into the source image space. Specifically, given a bounding box $b = [x_1, y_1, x_2, y_2]$, we generate a binary mask $\mathbf{M} \in \{0, 1\}^{H \times W}$, where pixels outside the bounding box are set to 1 and those within are set to 0. As illustrated in Fig.~\ref{fig:arch_fineedit}(a), We apply this mask to the source image $\mathbf{x}_{src}$ via element-wise multiplication, yielding a masked image $\mathbf{x}_{msk} = \mathbf{M} \odot \mathbf{x}_{src}$. Visually, $\mathbf{x}_{msk}$ preserves the visual information in the background while indicating the region of interest (ROI) through the zero-padded area.
\begin{align}
\mathbf{v} = f([[\mathbf{q}_{src} \parallel \mathbf 0], [\mathbf{\hat{q}}_{dst} \parallel \mathbf{q}_{msk}]] \mid t, c)
\end{align}
where $\mathbf{q}_{msk}$ denotes the latent representation of the masked image $\mathbf{x}_{msk}$. Here, the operators $[\cdot , \cdot]$ and $[\cdot \parallel \cdot]$ represent concatenation along the sequence and channel dimensions, respectively. The channel-wise fusion $[\mathbf{\hat{q}}_{dst} \parallel \mathbf{q}_{msk}]$ ensures that the denoising process is explicitly conditioned on both the preserved background context and the precise spatial localization of the editing area. For channel consistency, we concatenate the source latent $\mathbf{q}_{src}$ with a zero-padded tensor $\mathbf{0}$ to maintain identical channel dimensionality.


\subsubsection{Deep Fusion.} 
We employ a side adapter module to encode the masked latent into deep feature representations, which are subsequently injected into the intermediate layers of the diffusion transformer. Formally, for the $i$-th block, the output feature $\mathbf{h}_i$ is computed as:
\begin{align}
    \mathbf{h}_i = \mathcal{F}_i([[\mathbf{q}_{src} \parallel \mathbf 0], [\mathbf{\hat{q}}_{dst} \parallel \mathbf{q}_{msk}]] \mid t, c) + \mathcal{A}(\mathbf{q}_{msk})
\end{align}
where $\mathcal{F}_i$ denotes the backbone transformer block and $\mathcal{A}$ represents the side adapter. We utilize a 6-layer transformer as the side adapter, which introduces 10\% additional parameters relative to the backbone.
%
We observe that this module not only boosts localization capacity but also significantly accelerates convergence. This efficiency stems from the injection of source derived features directly into deeper layers, effectively functioning as a skip connection.

\begin{figure*}[t]
    \centering
    \includegraphics[width=1.0\linewidth]{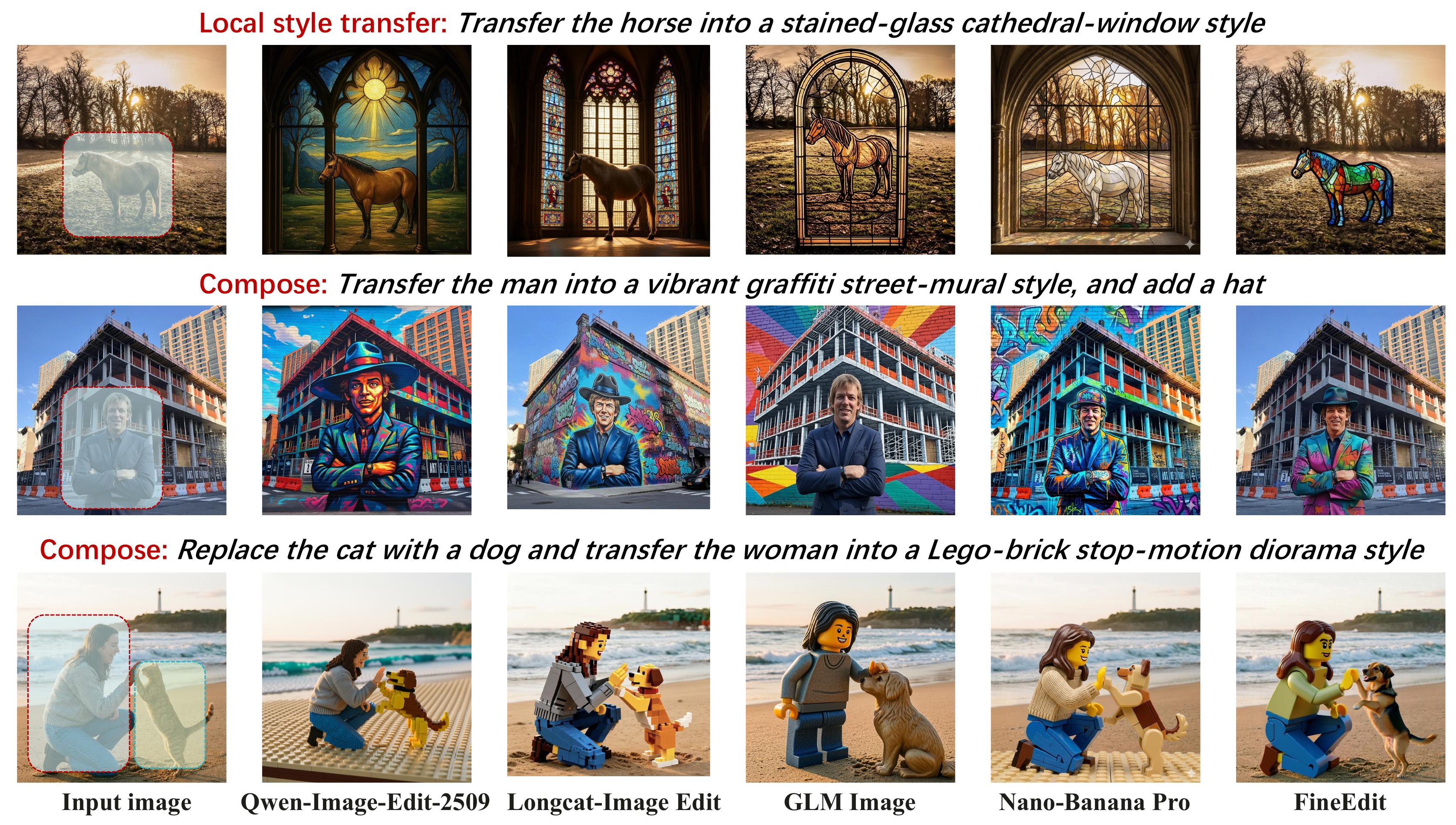}
        \vspace{-5mm}
    \caption{Unified visual instructions for diverse editing settings. Three representative configurations are showcased: localized style transfer, single-bbox composite editing, and multi-bbox composite editing. \name\ provides a flexible and consistent interface for both single-target and multi-target manipulation.}
    \label{fig:compose}
        \vspace{-4mm}
\end{figure*}

\subsection{Post-training: Reinforcement Learning with Decoupled Rewards}
Conventional reinforcement learning frameworks for image editing typically rely on a holistic reward function. These rewards, often derived from pre-trained VLMs such as GPT-4o, evaluate generated samples by computing global relative advantages. However, we contend that such coarse-grained signals are insufficient for fine-grained editing tasks, as they frequently fail to discern subtle semantic changes between the pre-edit and post-edit images. 

To address this, we leverage our proposed {\name-1.2M} dataset (detailed below), which provides dense bounding box annotations to decompose the original global reward into two complementary components: \textit{local editing fidelity} and \textit{background preservation}. Formally, the total reward $R(\mathbf{x}_{dst})$ is formulated as:
\begin{equation}
    R(\mathbf{x}_{dst}) = R_{roi} + R_{bg}
\end{equation}
Specifically, for the Region of Interest (ROI), we employ Qwen3-VL to compute $R_{roi}$, focusing on the semantic accuracy of the local edit. For the background reward $R_{bg}$, we integrate rule-based metrics with model-based scores to strike a balance between precise pixel-level preservation and perceptual naturalness:
\begin{align}
    R_{roi} &= \Phi_{\text{VLM}}(\mathbf{x}_{src} \odot \mathbf{M}, \mathbf{x}_{dst} \odot \mathbf{M}, c_1) \\
    R_{bg}  &= \Phi_{\text{VLM}}(\mathbf{x}_{src} \odot \bar{\mathbf{M}}, \mathbf{x}_{dst} \odot \bar{\mathbf{M}}, c_2) \\ \nonumber
    &+ \Psi_{\text{PSNR}}(\mathbf{x}_{src} \odot \bar{\mathbf{M}}, \mathbf{x}_{dst} \odot \bar{\mathbf{M}})
\end{align}

where $\mathbf{M}$ denotes the binary mask and $\bar{\mathbf{M}} = \mathbf{1} - \mathbf{M}$ represents its complement. The terms $c_1$ and $c_2$ are the corresponding VLM scoring prompts. $\Phi_{\text{VLM}}(\cdot)$ denotes a normalization function that maps the raw VLM scores to a bounded interval of $[0, 1]$, and $\Psi_{\text{PSNR}}(\cdot)$ represents the PSNR metric, which is further normalized via a min-max scaling approach.


\subsection{Unified Visual Instruction for Versatile Editing}
\label{sec:versatile_editing}

Beyond enhancing localization precision, our visual instruction mechanism serves as a unified interface that generalizes seamlessly across diverse editing granularity. As shown in Fig.~\ref{fig:compose}, by simply manipulating the spatial extent and quantity of the bounding boxes, \name\ can adapt to Local, Global, or Hybrid editing scenarios without architectural modifications~\footnote{More editing results are shown in the supplementary materials.}.

\noindent\textbf{Local and Global Unification.} 
The bounding box $B$ acts as a flexible controller for the editing scope. 
For standard \textit{Local Editing} tasks (e.g., Add, Replace, Remove), the user specifies a region $B_{local} \subset \mathcal{I}$ to constrain the generation within the object of interest. 
Conversely, for \textit{Global Editing} tasks such as global style transfer or background replacement, the instruction is naturally extended by expanding the bounding box to cover the entire image canvas ($B_{global} \approx \mathcal{I}$). This allows the model to leverage the same attention-masking mechanism to perform holistic regeneration while adhering to the text prompt.

\noindent\textbf{Hybrid Editing.} 
Our framework inherently supports multi-turn or multi-focus editing within a single pass. By providing multiple bounding boxes $\{B_1, B_2, \dots, B_n\}$, users can perform \textit{Hybrid Editing}, such as simultaneously replacing an object in the foreground while modifying a distant background element. The model attends to these disjoint regions independently, ensuring that complex, multi-objective instructions are executed without interference.

\begin{figure*}[t]
    \centering
    \includegraphics[width=1.0\linewidth]{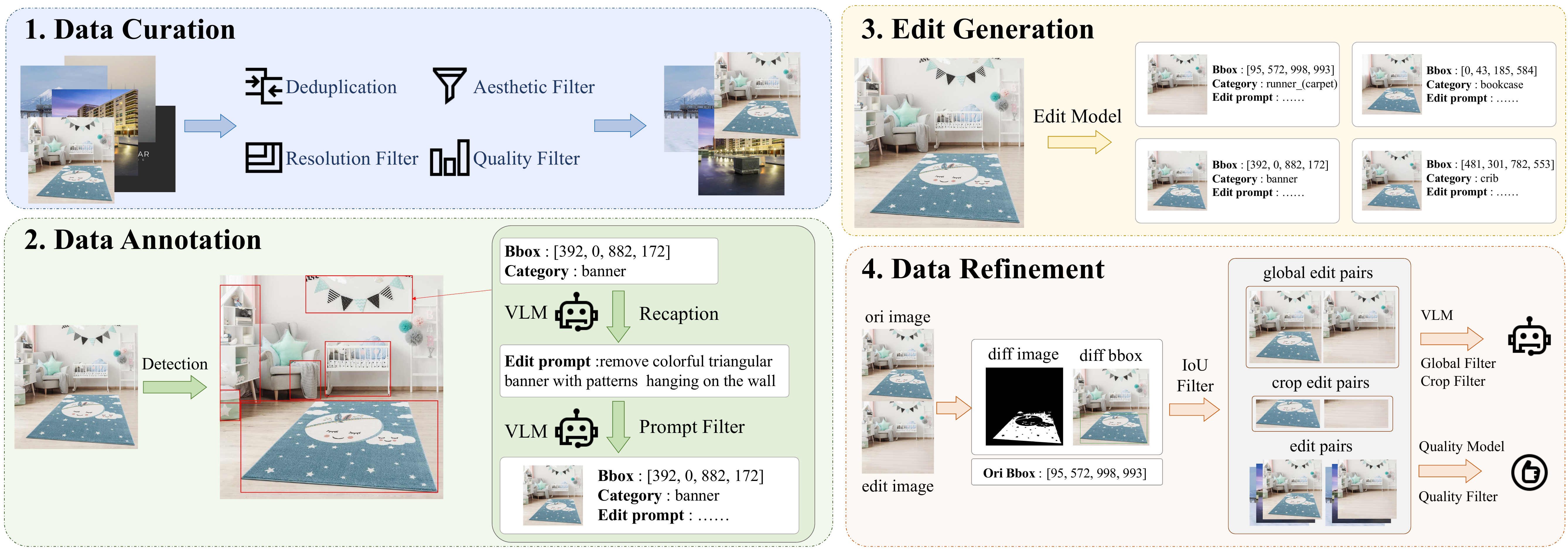}
    \vspace{-4mm}
    \caption{\textbf{The U-GAF Pipeline.}, which consists of four synergistic stages for high-quality data synthesis: 
\textbf{Data Curation, (2) Data Annotation, (3) Edit Generation, and (4) Data Refinement.}}
    \label{fig:data_pipe}
        \vspace{-4mm}
\end{figure*}

\noindent\textbf{Local Style Transfer.} 
Most notably, our experimental validation highlights a distinct capability of \name: \textit{Local Style Transfer}. 
Existing editing models often struggle to disentangle style from content spatially, typically applying style changes (e.g., ``oil painting style'', ``cyberpunk theme'') globally to the entire image. 
In contrast, by combining a local bounding box $B_{local}$ with a stylistic text prompt, our method effectively confines the style migration to the specific target object while preserving the semantic and stylistic integrity of the surrounding background. This granular control over style distribution represents a significant advancement over current SOTA methods.


\section{\name\ Dataset and Benchmark}
\label{sec:dataset}

Conventional approaches for constructing image editing datasets typically treat data pair generation and quality screening as decoupled stages. However, with the introduction of bounding box annotations to spatially localize editing regions, we observed that such a separated workflow leads to significant error accumulation across distinct steps. To address this challenge, we developed a unified pipeline (named U-GAF Pipeline) that integrates data generation, spatial bbox annotation, and rigorous data filtering into a cohesive framework. This synergistic approach ensures high alignment between visual content and annotations, serving as the foundation for our \name-1.2M dataset.

\begin{figure}[t]
    \centering
    \includegraphics[width=1.0\linewidth]{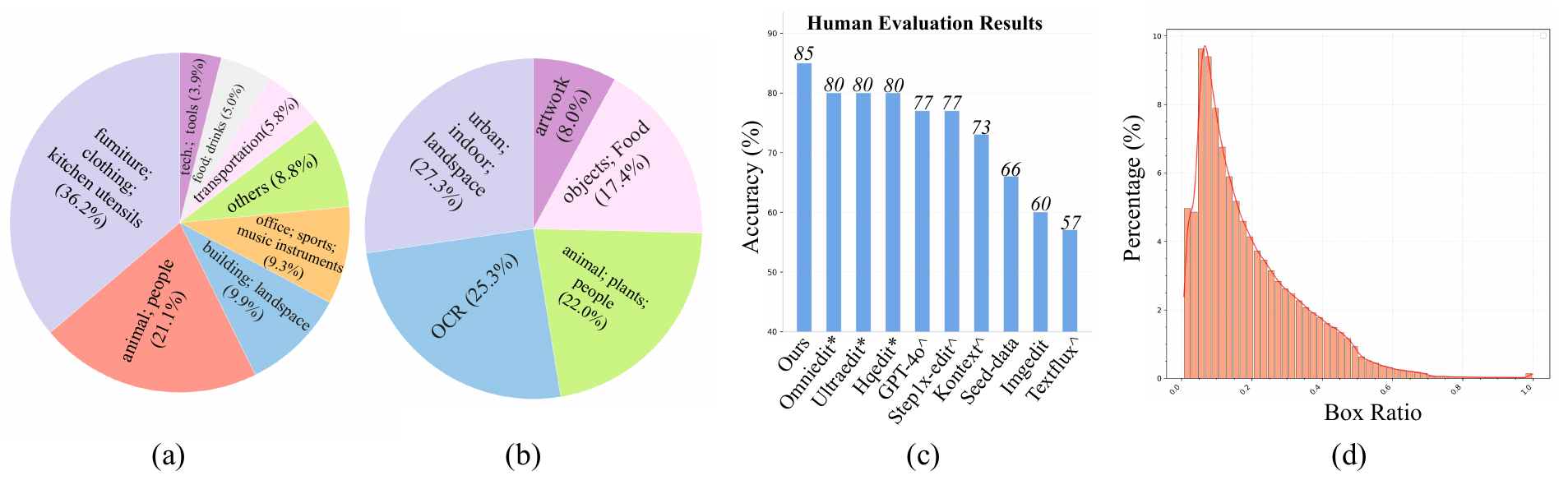}
    \caption{Information of our purposed FineEdit dataset. (a) Training dataset distribution. (b) Evaluation dataset distribution. (c) Human evaluation accuracy on different datasets. (d) Distribution of bounding box area ratios.}
    \label{fig:data_dist}
\end{figure}

\subsection{U-GAF Pipeline: Unified Generation, Annotation, and Filtering}

The proposed U-GAF Pipeline consists of four stages: 1)\textbf{Data Curation}, 2) \textbf{Data Annotation:} generation of bounding box locations and edit prompt, 3) \textbf{Edit Generation:} generation of edit pairs, and 4) \textbf{Data Refinement:} generation final image pairs. The flowchart of the Pipeline is illustrated in Fig.~\ref{fig:data_pipe}.

\noindent\textbf{Data Curation.} Our training data is curated from multiple sources, including Megalith-10M~\cite{matsubara2024megalith10m}, LAION-Aesthetic-6M~\cite{schuhmann2022laionaesthetic6plus}, and 5M internal images. To mitigate the bias towards simple scenes with sparse objects and synthetic datasets, we include diverse perception oriented datasets, including LVIS~\cite{DBLP:journals/corr/abs-1908-03195}, COCO~\cite{DBLP:journals/corr/LinMBHPRDZ14}, and CC3M~\cite{sharma-etal-2018-conceptual}. Following the filtering pipeline \cite{wu2025qwen} that excludes samples with suboptimal saturation, clarity, or low RGB entropy, we select a subset of 6M high-quality images.

%
%
%

\noindent\textbf{Data Annotation.}
We use Co-DETR~\cite{Zong_2023_ICCV} for object localization, and filter out roughly $20\%$ of the inital samples by excluding boxes with low confidence score (threshold 0.7) and extreme large box ratios. InternVL3-38B~\cite{zhu2025internvl3} is deployed to generate fine-grained editing instructions based on the refined visual inputs. Then we use Qwen3-32B-thinking to filter the editing instructions. The screening criteria aim to exclude the unreasonable image editing instructions, including inappropriate human-centric manipulations and content that violates safety guidelines.

%
%

\noindent\textbf{Edit Generation.} 
%
To generate diverse and high-quality edited samples, we incorporate multiple diffusion-based editing models as our synthetic engines, primarily Flux1-kontext-dev \cite{labs2025flux1kontextflowmatching} and Qwen-Image-Edit \cite{wu2025qwen}.

\noindent\textbf{Data Refinement.} To guarantee the high fidelity and precision of the FineEdit-1.2M dataset, we implement a rigorous four-stage sequential filtering pipeline, where each sample must successfully pass through every consecutive phase.
We first enforce spatial consistency by calculating the Intersection over Union (IoU) between the detected change mask (difference of original and edited images) and the ground-truth bounding box (generated by Co-DETR), discarding edits that drift significantly outside the target region.
Subsequently, we employ an RGB Entropy (RGBE) metric to assess information density, effectively filtering out trivial samples with low visual complexity (e.g., simple objects on monochromatic backgrounds). The pipeline then proceeds to semantic verification using Qwen3-VL-32B, which operates on two scales: a global pass to ensure overall visual coherence with the instruction, and a fine-grained local crop pass that specifically validates the editing quality within the designated bounding box. This coarse-to-fine strategy ensures that the final dataset maintains both global naturalness and precise local alignment with user prompts. Finally, we employ an image editing quality assessment model EditScore~\cite{luo2025editscoreunlockingonlinerl}, to filter out artifacts and enhance the realism of the final dataset.

Fig.~\ref{fig:data_dist}(a) and (d) provide a detailed breakdown of our \name-1.2M dataset. To assess the quality of \name-1.2M, we conduct a comprehensive evaluation involving both human annotation and quantitative comparisons with existing open-source image editing datasets, such as filtered Omniedit, Ultraedit, and Hqedit selected from GPT-Image-Edit-1.5M~\cite{wei2024omniedit,zhao2024ultraedit,hui2024hq,wang2025gptimageedit15mm}(denoted by ${}^*$ in Fig.~\ref{fig:data_dist}(c)), and filtered Kontext, Step1X-Edit, TextFlux, and GPT-4o selected from 
X2Edit Dataset~\cite{liu2025step1x,labs2025flux,ma2025x2edit} (denoted by \^{} in Fig.~\ref{fig:data_dist}(c)). As illustrated in the results, our dataset demonstrates superior annotation accuracy and broader stylistic diversity compared to prevailing alternatives. Specifically, human evaluators favor \name-1.2M for its precise alignment between textual instructions and spatial bounding-box cues, validating its effectiveness for training high-fidelity controllable editing models.

\subsection{\name-1k Evaluation Bench}
To build diversity in the evaluation dataset, we refer to the category settings used in the data collection of Qwen-Image~\cite{wu2025qwen}. We have set up data for ten categories: Objects, Urban Scenes, Indoors, Landscapes, Food, Plants, Animals, People, Text, and Artwork. 
Fig.~\ref{fig:data_dist}(b) shows the details of our \name-1k Evaluation Bench.
To ensure the high quality of our evaluation bench, we constructed three types of evaluation addition, removal, and replacement, based on the data from NHR-Edit~\cite{Layer2025NoHumansRequired} and MARIO-10M~\cite{chen2023textdiffuser} (Text data).
Specifically, we compute the pixel-wise difference between the pre- and post-edit images. Subsequently, morphological operations—including filtering, erosion, and dilation—are applied to generate a binary mask. Finally, the bounding box (bbox) is derived from the largest rectangular region within this mask.
To obtain high-quality prompts, we regenerate the `remove' prompts using Qwen2.5-7B and the `replace' prompts using Qwen2.5-VL-32B.
Finally, we conducted manual quality checks on all evaluation data pairs to ensure that the images were consistent with their categories, the prompts were appropriate, and the bounding box positions were correct.

\begin{table*}[!t]
    \centering
    \caption{Quantitative Comparison with State-of-the-Art methods on \name-1k Evaluation Bench. PC, VN, PDI, and OBR mean Prompt Compliance, Visual Naturalness, Physical \& Detail Integrity, and Out-of-Box Retention, respectively.}
    \label{tab:comparison_v2}
    \setlength{\tabcolsep}{3pt} 
    \resizebox{0.88 \linewidth}{!}{
    \begin{tabular}{l cccc cccc}
        \toprule
        & \multicolumn{4}{c}{\textbf{Background Preservation}} & \multicolumn{4}{c}{\textbf{BBox Editing Accuracy}} \\
        \cmidrule(lr){2-5} \cmidrule(lr){6-9}
        \textbf{Model} & \textbf{PSNR} $\uparrow$ & \textbf{SSIM}$\uparrow$ & \textbf{LPIPS}$\downarrow$ & \textbf{OBR}$\uparrow$  & \textbf{CLIP}$\uparrow$ & \textbf{PC}$\uparrow$ & \textbf{VN}$\uparrow$ & \textbf{PDI}$\uparrow$ \\
        \midrule
        Flux-fill-dev~\cite{flux2024}           & 34.2 & 0.90 & \textbf{0.08} & 4.99  & 0.26 & 3.56 & 4.03 & 3.98   \\
        BrushNet~\cite{ju2024brushnet}          & 29.6 & 0.83 & 0.13 & 4.99  & 0.25 & 3.28 & 3.38 & 3.34   \\
        BrushEdit~\cite{li2024brushedit}        & 30.9 & 0.84 & 0.13 & 4.95  & 0.25 & 3.27 & 3.45 & 3.41   \\
        PrefPaint~\cite{liu2024prefpaint}       & 27.1 & 0.77 & 0.21 & 4.83  & 0.23 & 2.74 & 3.01 & 2.96   \\
        Asuka-flux~\cite{wang2025towards}       & \textbf{34.4} & 0.88 & 0.16 & \textbf{5.00}  & 0.22 & 2.20 & 2.92 & 2.85   \\
        FluPA~\cite{fyp2025}                    & 32.7 & 0.89 & 0.09 & 4.96  & 0.26 & 3.68 & 4.05 & 4.01   \\
        \midrule
        GLM-Image~\cite{zhipuai2024glmimage}                & 27.4 & 0.85 & 0.14 & 4.48 & 0.25 & 3.66 & 3.98 & 3.95  \\
        Z-Image-turbo-Control~\cite{alibaba2025zimageturbo} & 28.7 & 0.86 & 0.10 & 4.99 & 0.24 & 3.18 & 3.51 & 3.48  \\
        LongCat-Image-Edit~\cite{LongCat-Image} & 27.3 & 0.83 & 0.12 & 4.38 & 0.26 & 4.59 & 4.52 & 4.56 \\
        Qwen-Image-Edit~\cite{wu2025qwen} &31.4 &0.88 &0.11 &4.23 &0.25 &3.94 & 4.21 &4.19\\
        Qwen-Image-Edit-2509~\cite{wu2025qwen} & 27.1 & 0.80 & 0.13 & 4.15 & 0.26 & 4.56 & 4.66 & 4.65  \\
        \midrule
        \textbf{\name} & \textbf{34.4} & \textbf{0.91} & \textbf{0.08} & 4.80  & \textbf{0.27} & \textbf{4.67} & \textbf{4.71} & \textbf{4.69}  \\
        \textbf{\name-r1} & \textbf{34.7} & \textbf{0.91} & \textbf{0.09} & 4.89 & \textbf{0.27} & \textbf{4.80} & \textbf{4.83} & \textbf{4.81}  \\
        \bottomrule
    \end{tabular}
    }
\end{table*}

\section{Experiments}
\label{sec:exp}
\subsection{Experimental Setup}
\textbf{Implementation Details.} 
The training of \name\ follows a two-stage paradigm: pre-training followed by reinforcement learning-based post-training. 
in pre-training Stage, we train \name\ on the \name-1.2M dataset using 64 NVIDIA H100 GPUs. The per-device batch size is set to 1, with gradient accumulation over 8 steps to achieve an effective global batch size of 512. We employ the AdamW optimizer with a learning rate of $5 \times 10^{-5}$. All training images are resized to a fixed resolution of $1024 \times 1024$, and the total training duration is 15,000 steps. 
in post-training stage, following standard RL post-training protocols~\cite{zheng2025diffusionnft}, we incorporate LoRA~\cite{hulora} layers with a rank of $r=32$ and an alpha of $\alpha=64$, while keeping the transformer backbone frozen. The learning rate for this stage is $3 \times 10^{-4}$. The RL optimization is organized into 24 groups per epoch, with a group size of 16 for relative advantage calculation.


\textbf{Evaluation Benchmarks.} 
We evaluate the performance of our method on the \name\-1k Benchmark, a curated test set designed to assess fine-grained editing capabilities. To further demonstrate the generalizability of \name, we also report results on the {Single-turn} subset of ImgEdit-Bench~\cite{ye2025imgedit} and GEdit-Bench-EN~\cite{liu2025step1x}, respectively. These two test set encompasses a broader range of editing categories beyond the standard addition, removal, and replacement tasks, providing a robust test for versatile editing scenarios.

\textbf{Evaluation Metrics.} 
To comprehensively assess performance, we employ a multi-dimensional evaluation protocol covering reconstruction quality, semantic alignment, and perceptual fidelity:
\textbf{Low-level Metrics:} We report PSNR, SSIM, and LPIPS calculated specifically on regions outside the boxes to quantify background preservation. \textbf{Semantic Alignment:} We utilize CLIP scores to measure the correspondence between edited results and text instructions within the bbox. \textbf{VLM Evaluation:} We use VLM to evaluate on four dimensions~\cite{ye2025imgedit}: {Prompt Compliance} (PC), {Visual Naturalness} (VN), {Physical \& Detail Integrity} (PDI), and {Out-of-Box Retention} (OBR). Notably, OBR assesses the semantic consistency of non-edited regions to complement pixel-level metrics.

\subsection{Quantitative Results}
We evaluate \name\ against SOTA methods on three benchmarks:  \name-1k bench, ImgEdit-bench and GEdit-bench. All inference results are generated using a classifier-free guidance scale of $4$ and $30$ sampling steps.

\noindent\textbf{Performance on \name-1k Bench.} 
In Table~\ref{tab:comparison_v2}, we categorize the baseline models into two groups based on their input modalities. The upper section comprises Flux-based models~\cite{labs2025flux}, which typically generate edited results via a mask and a descriptive prompt. Since their prompt requirements differ from instruction-based methods (e.g., add/remove...), we manually refined the prompts for these baselines to better align with their architectural priors. This adaptation yielded a significant performance boost, ensuring a fair and rigorous comparison. The lower section presents recent specialized editing methods that utilize direct bounding box inputs for spatial localization. As evidenced by the results, \name\ consistently outperforms all competing methods. Notably, \name\ achieves superior scores in both region-of-interest (ROI) editing fidelity and background preservation, demonstrating its robust localization and high-fidelity synthesis capabilities. Furthermore, our model variant subjected to RL post-training, denoted as \name-r1, yields further performance gains.

\begin{table*}[t!]
\centering
\caption{Quantitative Comparison with SOTA methods on ImgEdit-Bench.}
 \label{tab:comparison_imgedit}
\resizebox{\linewidth}{!}{
\begin{tabular}{l|cccccccccc}
\hline
\textbf{Model} & \textbf{Add} & \textbf{Adjust} & \textbf{Extract} &\textbf{Replace} & \textbf{Remove} & \textbf{Background} & \textbf{Style} & \textbf{Hybrid} & \textbf{Action} & \textbf{Overall}\\ \hline
OmniGen~\cite{xiao2025omnigen}                  & 3.47 & 3.04 & 1.71 &2.94 & 2.43 & 3.21 & 4.19 & 2.24 & 3.38 & 2.96\\
ICEdit~\cite{zhang2025context}                  & 3.58 & 3.39 & 1.73 &3.15 & 2.93 & 3.08 & 3.84 & 2.04 & 3.68 & 3.05\\
Step1X-Edit~\cite{liu2025step1x}                & 3.88 & 3.14 & 1.76 &3.40 & 2.41 & 3.16 & 4.63 & 2.64 & 2.52 & 3.06\\
BAGEL~\cite{deng2025emerging}                   & 3.56 & 3.31 & 1.70 &3.30 & 2.62 & 3.24 & 4.49 & 2.38 & 4.17 & 3.20\\
UniWorld-V1~\cite{lin2025uniworld}              & 3.82 & 3.64 & 2.27 &3.47 & 3.24 & 2.99 & 4.21 & 2.96 & 2.74 & 3.26\\
OmniGen2~\cite{wu2025omnigen2}                  & 3.57 & 3.06 & 1.77 &3.74 & 3.20 & 3.57 & 4.81 & 2.52 & 4.68 & 3.44\\
Kontext-pro~\cite{labs2025flux}                 & 4.25 & 4.15 & 2.35 &4.56 & 3.57 & 4.26 & 4.57 & 3.68 & 4.63 & 4.00\\
Kontext-dev~\cite{labs2025flux}                 & 4.12 & 3.80 & 2.04 &4.22 & 3.09 & 3.97 & 4.51 & 3.35 & 4.25 & 3.71\\
GPT-4o-Image~\cite{OpenAI2025Image}             & \textbf{4.61} & 4.33 & 2.90 &4.35 & 3.66 & \textbf{4.57} & \textbf{4.93} & \textbf{3.96} & \textbf{4.89} & 4.20\\
Qwen-Image-Edit~\cite{wu2025qwen}               & 4.38 & 4.16 & 3.43 &4.66 & 4.14 & 4.38 & 4.81 & 3.82 & 4.69 & 4.27\\
Qwen-Image-Edit-2509~\cite{wu2025qwen}          & 4.32 & 4.36 & \textbf{4.04} &4.64 & 4.52 & 4.37 & 4.84 & 3.39 & 4.71 & 4.35\\
\midrule
\textbf{\name}                                   & 4.53 & \textbf{4.57} & 4.03 &\textbf{4.83} & \textbf{4.65} & 4.53 & 4.68 & 3.85 & 4.86 & \textbf{4.50} \\
\bottomrule
\end{tabular}
}
\end{table*}

\begin{table*}[t!]
\centering
\caption{Quantitative Comparison with SOTA methods on GEdit-Bench-EN.}
\vspace{-2mm}
\label{tab:gedit_bench}
\resizebox{0.96\linewidth}{!}{
\begin{tabular}{lccc|lccc}
\toprule
\textbf{Model} & \textbf{G\_SC} & \textbf{G\_PQ} & \textbf{G\_O} & \textbf{Model} & \textbf{G\_SC} & \textbf{G\_PQ} & \textbf{G\_O} \\ 
\midrule

Instruct-P2P~\cite{brooks2023instructpix2pix} & 3.58 & 5.49 & 3.68 & GPT-Image-1 [High] & 7.85 & 7.62 & 7.53 \\
AnyEdit~\cite{yu2025anyedit}   & 3.18 & 5.82 & 3.21 & Gemini 2.0 & 6.73 & 6.61 & 6.32 \\
MagicBrush~\cite{zhang2023magicbrush} & 4.68 & 5.66 & 4.52 & Kontext-dev~\cite{labs2025flux} & 6.52 & 7.38 & 6.00 \\
UniWorld-V1~\cite{lin2025uniworld}  & 4.93 & 7.43 & 4.85 & Kontext-pro~\cite{labs2025flux} & 7.02 & 7.60 & 6.56\\
OmniGen~\cite{xiao2025omnigen}  & 5.96 & 5.89 & 5.06 & UniPic2 ~\cite{wei2025skyworkunipic20}& - & - & 7.10 \\
OmniGen2~\cite{wu2025omnigen2}  & 7.16 & 6.77 & 6.41 & Qwen-Image-Edit~\cite{wu2025qwen}  & 8.00 & 7.86 & 7.56 \\
BAGEL~\cite{deng2025emerging}  & 7.36 & 6.83 & 6.52  & {Qwen-Image-Edit-2509}~\cite{wu2025qwen}  & 8.15 & \textbf{7.86} & 7.54\\
Step1X-Edit~\cite{liu2025step1x}   & 7.66 & 7.35 & 6.97  &\cellcolor{gray!15}{\name} & \cellcolor{gray!15}{\textbf{8.71}} &\cellcolor{gray!15}{7.47} &\cellcolor{gray!15}{\textbf{7.97}} \\
\bottomrule
\end{tabular}%
}
\end{table*}

%

\begin{figure*}[t]
    \centering
    \includegraphics[width=1.0\linewidth]{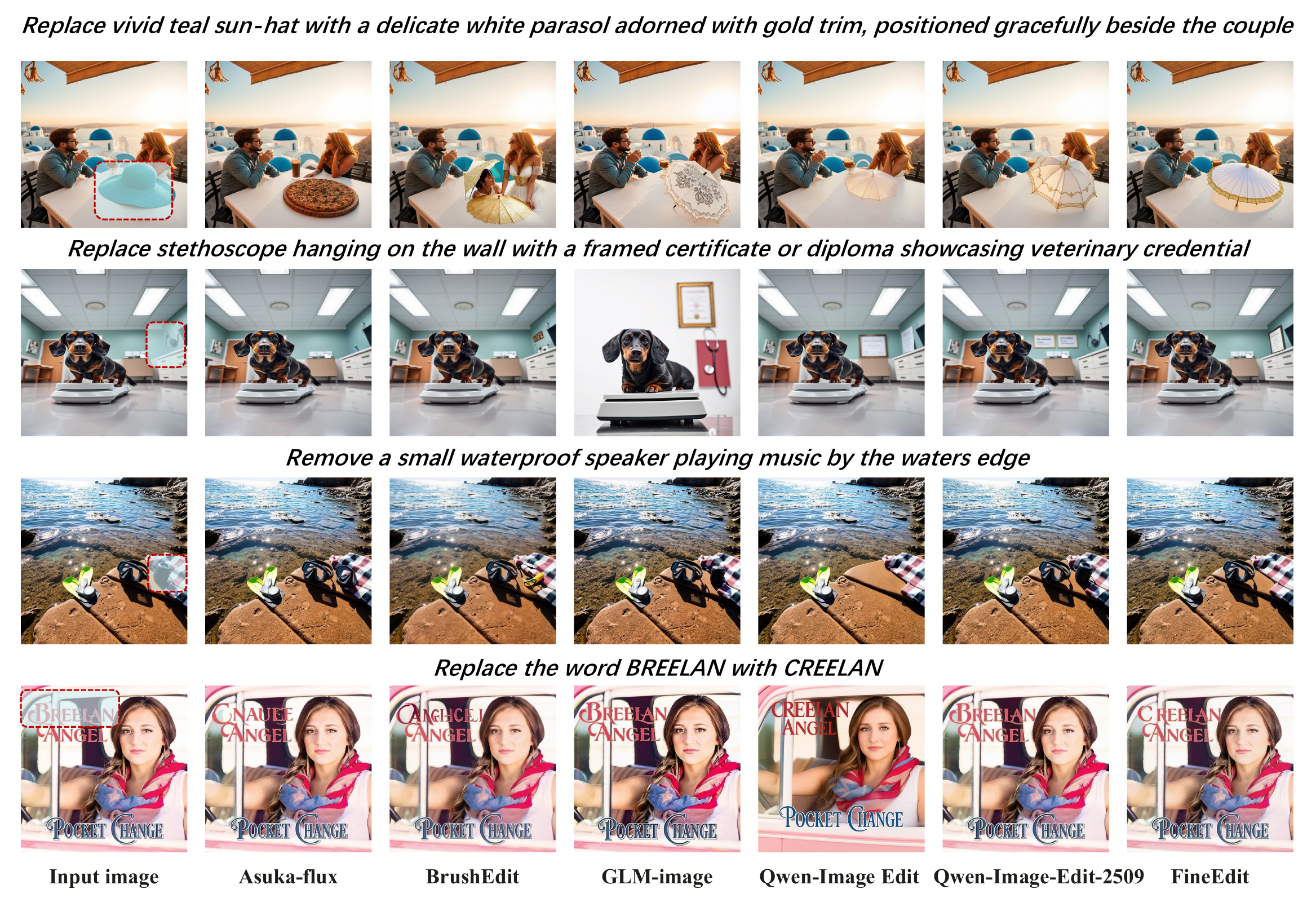}
        \vspace{-6mm}
    \caption{Comparision on \name-1k Evaluation Bench.}
    \label{fig:visual_results_bbedit_1k}
        \vspace{-4mm}
\end{figure*}

\begin{table*}[t]
\centering
\caption{Ablation study on fusion strategies and reward functions.}
\vspace{-2mm}
\label{tab:ablation_study}
\resizebox{0.92\linewidth}{!}{%
\begin{tabular}{llcccc|cccc}
\toprule
\multirow{2}{*}{\textbf{Ablation}} & \multirow{2}{*}{\textbf{Module}} & \multicolumn{4}{c|}{\textbf{Background Preservation}} & \multicolumn{4}{c}{\textbf{Bbox Editing Accuracy}} \\
\cmidrule(lr){3-6} \cmidrule(lr){7-10}
& & \textbf{PSNR} $\uparrow$ & \textbf{SSIM} $\uparrow$ & \textbf{LPIPS} $\downarrow$ & \textbf{OBR} $\uparrow$ & \textbf{CLIP} $\uparrow$ & \textbf{PC} $\uparrow$ & \textbf{VN} $\uparrow$ & \textbf{PDI} $\uparrow$ \\
\midrule
\multirow{4}{*}{Fusion} 
& Baseline~\cite{wu2025qwen} & 27.1 & 0.80 & 0.13 & 4.15 & 0.26 & 4.56 & 4.66 & 4.65 \\
& Early & 15.9 & 0.48 & 0.40 & 2.43 & 0.21 & 2.45 & 1.85 & 1.85 \\
& Deep & 33.1 & 0.88 & 0.10 & 4.61 & 0.26 & 4.52 & 4.51 & 4.61 \\
& Early + Deep & 34.4 & 0.91 & 0.08 & 4.80 & 0.27 & 4.67 & 4.71 & 4.69 \\
\midrule
\multirow{3}{*}{Reward} & Pre-RL & 34.4 & 0.91 & 0.08 & 4.80 & 0.27 & 4.67 & 4.71 & 4.69 \\
& Global & 32.7 & 0.90 & 0.10 & 4.76 & 0.27 & 4.74 & 4.76 & 4.73 \\
& Decoupled & {34.7} &{0.91} & {0.09} & {4.89} & {0.27} & {4.80} & {4.83} & {4.81} \\
\bottomrule
\end{tabular}}
\end{table*}

\noindent\textbf{Performance on ImgEdit and GEdit Bench.} 
In Table~\ref{tab:comparison_imgedit} and Table~\ref{tab:gedit_bench}, we further demonstrate the versatility of our approach. Beyond the standard \textit{add}, \textit{remove}, and \textit{replace} tasks, \name\ maintains robust performance across a diverse range of editing scenarios by simply adjusting the bounding box format. This underscores the flexibility of our unified visual instruction in handling complex, multi-target, and out-of-distribution editing tasks. For fair comparisons, we report results obtained solely after the pre-training stage, without the task specific RL fine-tuning across both evaluation benchmarks.

\subsection{Qualitative Results}
In Fig.~\ref{fig:visual_results_bbedit_1k}, we present a visual comparison of our method with other state-of-the-art methods.
The red boxes in the first column indicate the regions to be edited. The second example is a case of complex prompt coupled with fine-grained object replacement. Other methods produced editing errors (Asuka-flux removed the object but did not add the certificate; BrushNet added the wrong object; Qwen-Image-Edit added objects outside the red box; And GLM-image incorrectly extracted the object).
The last example demonstrates the text editing capability. Our method successfully modified the word while matching the color and font style with the original input image.
Fig.~\ref{fig:compose} demonstrates that our method exhibits superior local editing capabilities (as seen in the first and second samples) and hybrid editing (third sample) compared to existing approaches.
More results can be found in the supplementary materials.

\vspace{-1mm}
\subsection{Ablation Study}
\noindent\textbf{Impact of Early and Deep Fusion.} 
We incorporate bounding box conditions into \name\ using both early and deep fusion mechanisms. To evaluate their individual contributions, we conduct an ablation study isolating each strategy. As shown at the top of Table~\ref{tab:ablation_study}, combining both methods yields optimal performance. Specifically, using early fusion alone causes convergence issues, despite initialization with pre-trained Qwen-Image-Edit weights. Conversely, deep fusion alone avoids these issues by preserving the original input manifold at the initial layer, yet it still underperforms the full model. This demonstrates that early fusion is essential for providing strong spatial priors, while deep fusion ensures robust multi-level feature alignment.

\noindent\textbf{Impact of Decoupled Reward for RL.} 
We investigate the influence of our proposed decoupled reward function during the post-training stage. As summarized in the bottom of Table~\ref{tab:ablation_study}, we compare its performance against a conventional global reward mechanism. Our results indicate that while the global reward yields only marginal improvements in localized editing accuracy, it significantly compromises background preservation. In contrast, by utilizing the decoupled reward function, both editing precision within the bounding box and background fidelity are consistently enhanced. Notably, we observe a substantial boost in region-specific editing accuracy, validating the effectiveness of our strategy in balancing local manipulation with global context retention.

\section{Conclusion}
This paper present \name, a framework designed to address the challenges of precise spatial localization and background consistency in diffusion-based image editing. By shifting from text-only prompts to multi-level bounding-box guidance, \name\ enables intuitive, high precision control over editing targets. To facilitate the training and evaluation of these capabilities, we introduce \name-1.2M, a dataset comprising 1.2 million annotated pairs, alongside \name-Bench, a comprehensive benchmark for region-based editing. Extensive experiments demonstrate that our approach significantly outperforms state-of-the-art models in instruction compliance and perceptual fidelity. Furthermore, \name\ generalizes robustly across diverse edit benchmarks. This unified visual instruction paradigm, together with the accompanying dataset, provides a strong foundation for future research in high-fidelity, controllable image editing.

\noindent\textbf{Limitations and Future Work.} Although \name\ excels in spatial control and background preservation, it relies primarily on bounding boxes for visual guidance. While effective for defining object boundaries, bounding boxes lack the flexibility needed for fine-grained tasks like irregular shape modifications or precise path-based editing. Future work will extend \name\ to support interactive prompts such as points, scribbles, and polygons, enabling more granular control over generation.

\bibliographystyle{splncs04}
\bibliography{main}

\clearpage

\section*{Supplementary Material}

\input{appendix}
\end{document}

%% file: appendix.tex
In the supplementary material, we first provide additional qualitative results for \name\ and extend our comparisons with state-of-the-art editing models. We further evaluate \name\ on the \name-1k benchmark against a broader range of baselines. Next, we visualize the performance of \name\ across various bounding box scales and assess its robustness to spatial perturbations in box placement, simulating potential human interaction errors. Finally, we present human evaluation results comparing \name\ against Qwen-Image-Edit~\cite{wu2025qwen}, Qwen-Image-Edit-2509~\cite{wu2025qwen}, and Longcat-Image-Edit~\cite{LongCat-Image}.

\section{Extended Qualitative Results}

\subsection{Qualitative Results on \name-1k}
\begin{figure*}[t]
    \centering
    \includegraphics[width=0.95\linewidth]{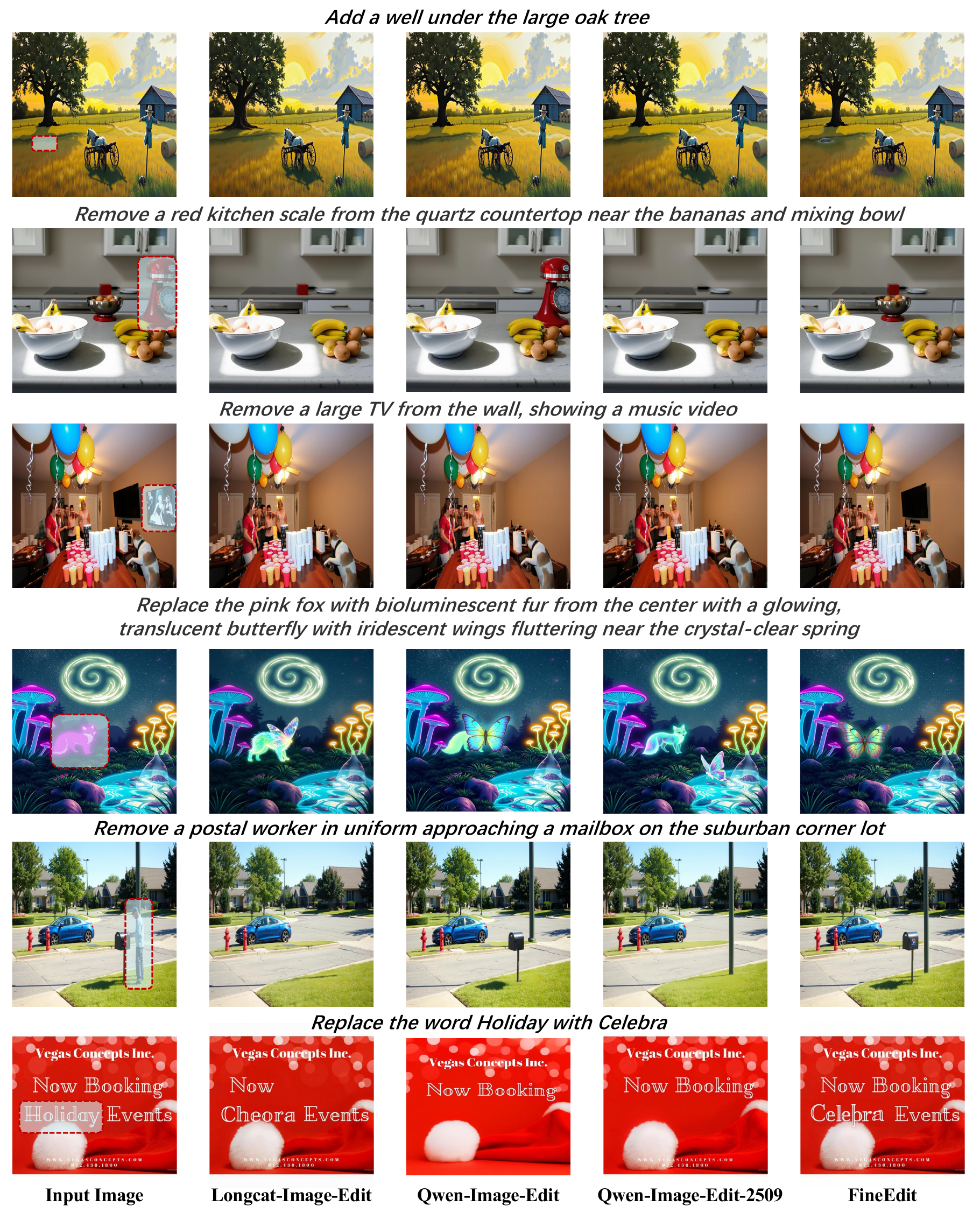}
    \caption{Extended Qualitative Results on FineEdit-1k}
    \label{fig:more-nhr-1}
\end{figure*}

While Figure 6 in the main text provides initial visualizations on the \name-1k benchmark, this section offers a more comprehensive suite of results. Due to space constraints in the main manuscript, we present additional qualitative comparisons between \name, Qwen-Image-Edit, Qwen-Image-Edit-2509 and LongCat-Image-Edit here. These extended visual results, illustrated in Fig.~\ref{fig:more-nhr-1}, further demonstrate our model's performance across diverse scenarios.

\subsection{Qualitative Results Across Diverse Editing Tasks}
\begin{figure}[t]
    \centering
    \includegraphics[width=1.0\linewidth]{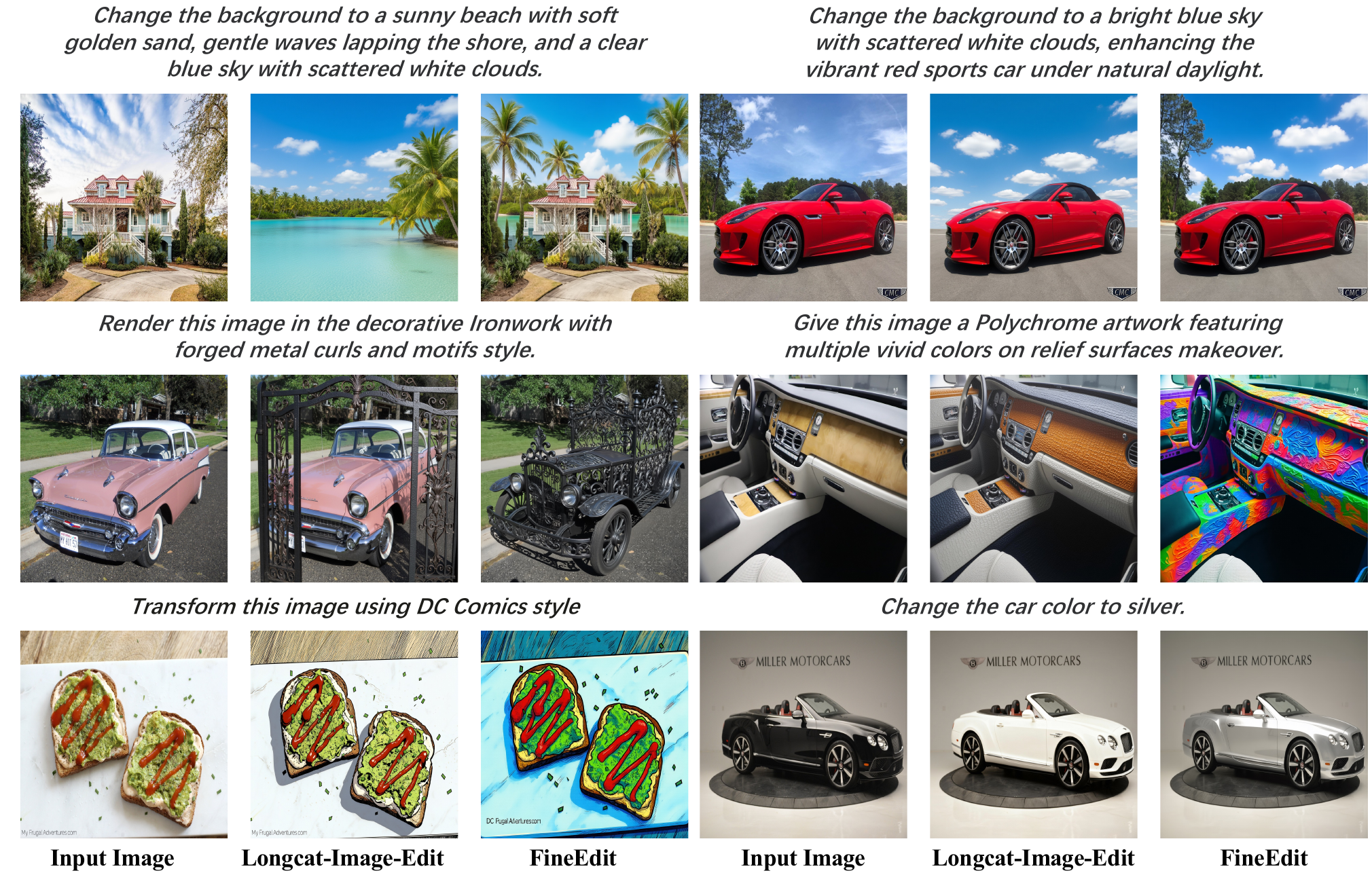}
    \caption{Qualitative Results Across Diverse Editing Tasks}
    \label{fig:more-task}
\end{figure}

Following the qualitative analysis of local style transfer, object addition, removal, and replacement presented in the main manuscript (Fig. 3 and 6), we provide additional visualizations here. Fig.~\ref{fig:more-task} illustrates these extended results and offers a comparative study against Longcat-Image-Edit, further validating the efficacy of \name\ in general editing scenarios.

\begin{figure}[t]
    \centering
    \includegraphics[width=1.0\linewidth]{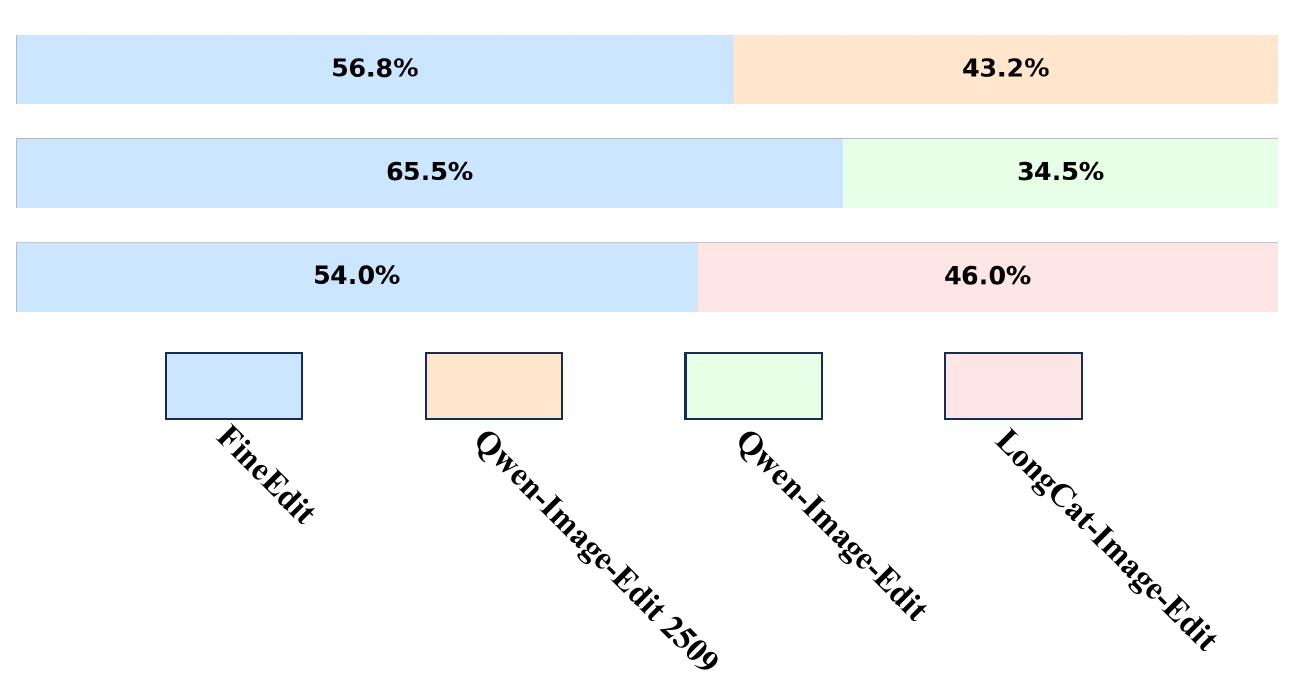}
    \vspace{-5mm}
    \caption{Comparison of human evaluation win rates between \name\ and competing methods.}
    \label{fig:human-eval}
    \vspace{-4mm}    
\end{figure}

\section{Extended Evaluation on the \name-1k Benchmark}
While Table 1 in the main manuscript primarily compares \name\ with Flux-based methods and several recently released open-source SOTA models, this section provides a broader evaluation on the \name-1k benchmark. Specifically, we include additional comparisons with classical image editing baselines and the recent FireRed-Image-Edit~\cite{superintelligenceteam2026fireredimageedit10technicalreport}. As summarized in Table~\ref{tab:comparison_v3}, \name\ consistently outperforms all compared methods, demonstrating its superior performance.

\section{Robustness to Bounding Box Variations}
In this section, we evaluate the robustness of \name\ with respect to variations in bounding box (bbox) specifications. Specifically, we first analyze the model's editing performance across a wide range of bbox scales. Furthermore, we conduct a sensitivity analysis by introducing spatial perturbations to the bbox coordinates to simulate potential inaccuracies in human interaction.

\subsection{Robustness to Diverse Bounding Box Scales}
\begin{figure}[t]
    \centering
    \includegraphics[width=1.0\linewidth]{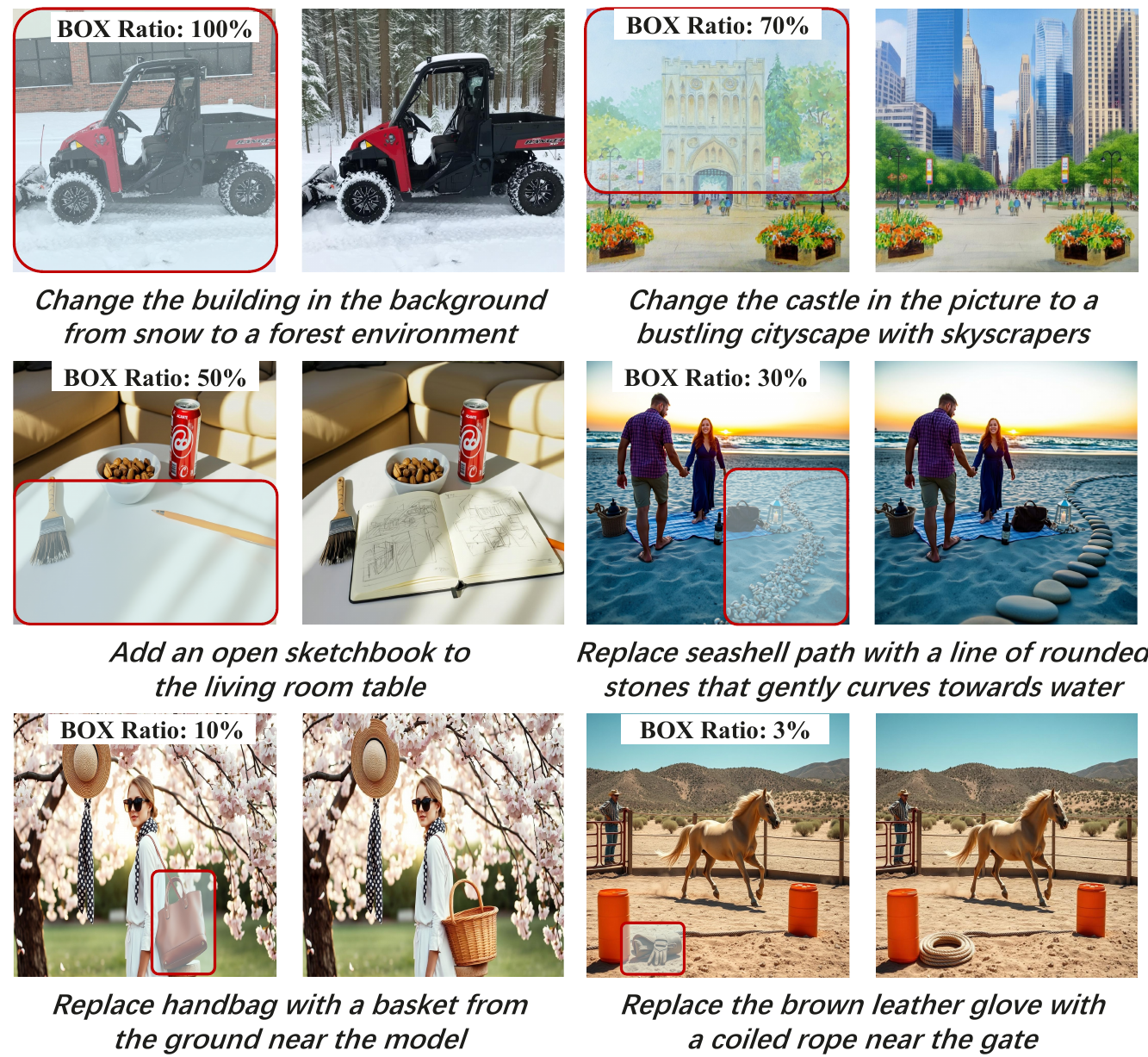}
    \vspace{-5mm}
    \caption{Robustness to Various Bounding Box Scales and Ratios.}
    \label{fig:rob-bbox-ratio}
    \vspace{-4mm}    
\end{figure}

As illustrated in Fig.~\ref{fig:rob-bbox-ratio}, \name\ demonstrates consistent performance across diverse bbox scales and ratios, maintaining high-quality editing results from expansive regions to extremely localized areas.

\begin{table*}[!t]
    \centering
    \caption{Quantitative Comparison with State-of-the-Art methods. PC, VN, PDI, and OBR mean Prompt Compliance, Visual Naturalness, Physical \& Detail Integrity, and Out-of-Box Retention, respectively.}
    \label{tab:comparison_v3}
    \setlength{\tabcolsep}{3pt} 
    \resizebox{0.95 \linewidth}{!}{
    \begin{tabular}{l cccc cccc}
        \toprule
        & \multicolumn{4}{c}{\textbf{Background Preservation}} & \multicolumn{4}{c}{\textbf{BBox Editing Accuracy}} \\
        \cmidrule(lr){2-5} \cmidrule(lr){6-9}
        \textbf{Model} & \textbf{PSNR} $\uparrow$ & \textbf{SSIM}$\uparrow$ & \textbf{LPIPS}$\downarrow$ & \textbf{OBR}$\uparrow$  & \textbf{CLIP}$\uparrow$ & \textbf{PC}$\uparrow$ & \textbf{VN}$\uparrow$ & \textbf{PDI}$\uparrow$ \\
        \midrule
        FireRed-Image-Edit~~\cite{superintelligenceteam2026fireredimageedit10technicalreport}     & 24.2 & 0.80 & 0.16 & 4.06 & 0.35 & 4.48 & 4.52 & 4.51 \\
        Bagel~\cite{deng2025emerging}                  & 29.3 & 0.87 & 0.14 & 4.11 & 0.35 & 4.06 & 3.95 & 3.93 \\
        Stepx-edit-v1p2~\cite{liu2025step1x}        & 27.9 & 0.83 & 0.13 & 3.98 & 0.36 & 4.47 & 4.50 & 4.50 \\
        OmniGen2  ~\cite{wu2025omnigen2}              & 25.1 & 0.83 & 0.19 & 3.32 & 0.34 & 3.60 & 3.69 & 3.67 \\
        \midrule
        \textbf{\name}         & \textbf{34.4} & \textbf{0.91} & \textbf{0.08} & \textbf{4.80}  & \textbf{0.27} & \textbf{4.67} & \textbf{4.71} & \textbf{4.69}  \\
        \textbf{\name-r1}      & \textbf{34.7} & \textbf{0.91} & \textbf{0.09} & \textbf{4.89}  & \textbf{0.27} & \textbf{4.80} & \textbf{4.83} & \textbf{4.81}  \\
        \bottomrule
    \end{tabular}
    }
    \vspace{-2mm}
\end{table*}

\subsection{Robustness to Bounding Box Perturbations}
\begin{figure}[t]
    \centering
    \includegraphics[width=1.0\linewidth]{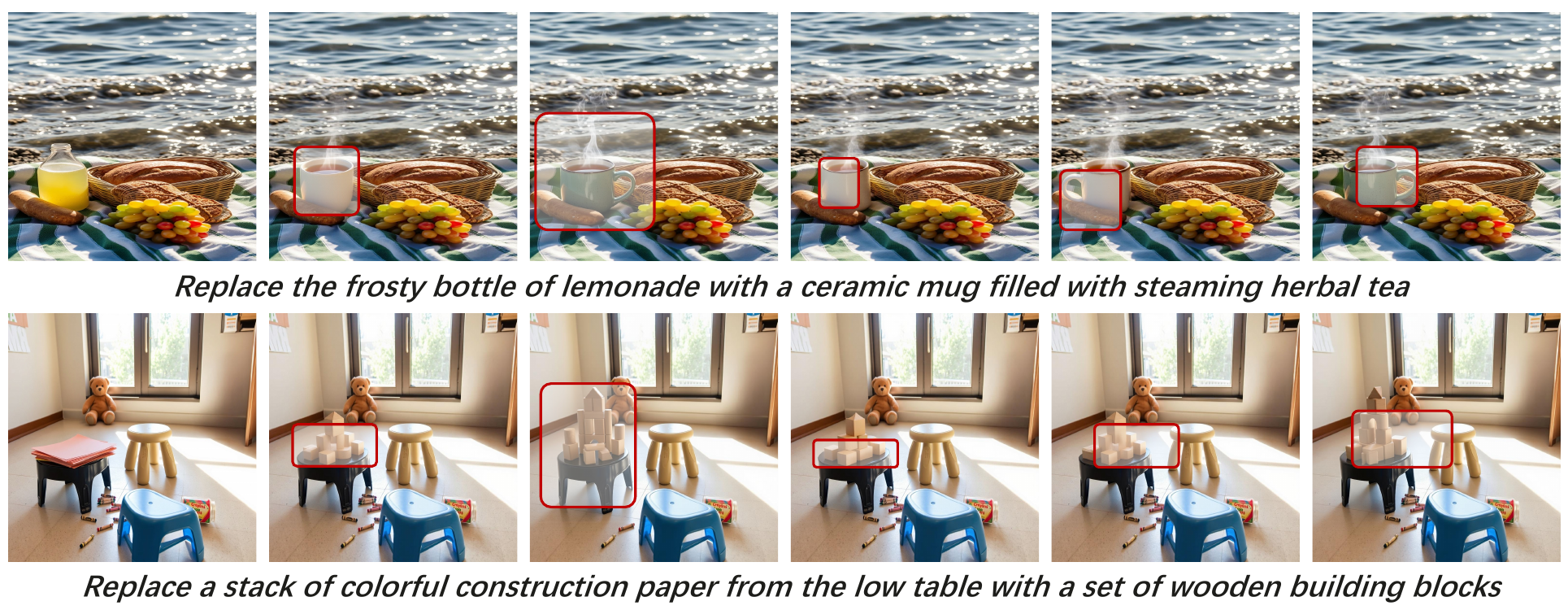}
    \vspace{-5mm}
    \caption{Robustness to Bounding Box Perturbations.}
    \label{fig:rob-bbox-pert}
    \vspace{-4mm}    
\end{figure}

As illustrated in Fig.~\ref{fig:rob-bbox-pert}, \name\ maintains high precision editing capabilities even when provided with imprecise or noisy bounding box instructions. This robustness to spatial jitter demonstrates the model's reliability in practical scenarios where human-provided guidance may be coarse or misaligned.

\section{Human Evaluation}
We conducted a human evaluation to perform a side-by-side comparison between \name\ and several competitive baselines, including Qwen-Image-Edit~\cite{wu2025qwen}, Qwen-Image-Edit-2509, and Longcat-Image-Edit~\cite{LongCat-Image}.

Specifically, we conducted independent blind trials on the \name-1k benchmark, involving ten human volunteers for the evaluation. To ensure an objective assessment and eliminate potential bias, the identities of the models were kept hidden from the evaluators. They were instructed to select the superior result based on a comprehensive evaluation of semantic consistency and visual quality. As shown in the preference scores in Fig.~\ref{fig:human-eval}, our method demonstrates a clear advantage over state-of-the-art models.

